\documentclass[sigconf]{acmart}
\usepackage{blindtext}
\usepackage{balance}

\AtBeginDocument{%
  \providecommand\BibTeX{{%
    \normalfont B\kern-0.5em{\scshape i\kern-0.25em b}\kern-0.8em\TeX}}}

\copyrightyear{2021} 
\acmYear{2021} 
\setcopyright{acmcopyright}\acmConference[ICMR '21]{Proceedings of the 2021 International Conference on Multimedia Retrieval}{August 21--24, 2021}{Taipei, Taiwan}
\acmBooktitle{Proceedings of the 2021 International Conference on Multimedia Retrieval (ICMR '21), August 21--24, 2021, Taipei, Taiwan}
\acmPrice{15.00}
\acmDOI{10.1145/3460426.3463655}
\acmISBN{978-1-4503-8463-6/21/08}

\usepackage{soul}
\usepackage[colorinlistoftodos,prependcaption,textsize=tiny]{todonotes}

\definecolor{dynamic}{RGB}{245, 186, 64}
\definecolor{static}{RGB}{116, 113, 174}

\usepackage{colortbl}

\newcommand{\Gc}{\cellcolor{gray!45}}

\def\etal{\emph{et al}.}

\begin{document}

\fancyhead{}

\title{Scene-aware Learning Network for Radar Object Detection}

\author{Zangwei Zheng$^{1}$\quad Xiangyu Yue$^{2}$\quad Kurt Keutzer$^{2}$\quad Alberto Sangiovanni Vincentelli$^{2}$ \\
$^1$Nanjing University\quad $^2$UC Berkeley
}




\renewcommand{\shortauthors}{Zangwei, \etal}

\begin{abstract}
  Object detection is essential to safe autonomous or assisted driving. 
  Previous works usually utilize RGB images or LiDAR point clouds to identify and localize multiple objects in self-driving. However, cameras tend to fail in bad driving conditions, \textit{e.g.} bad weather or weak lighting, while LiDAR scanners are too expensive to get widely deployed in commercial applications. Radar has been drawing more and more attention due to its robustness and low cost.  
  In this paper, we propose a scene-aware radar learning framework for accurate and robust object detection. First, the learning framework contains branches conditioning on the scene category of the radar sequence; with each branch optimized for a specific type of scene. 
  Second, three different 3D autoencoder-based architectures are proposed for radar object detection and ensemble learning is performed over the different architectures to further boost the final performance.
  Third, we propose novel scene-aware sequence mix augmentation (SceneMix) and scene-specific post-processing to generate more robust detection results. In the ROD2021 Challenge, we achieved a final result of average precision of 75.0\% and an average recall of 81.0\%. Moreover, in the parking lot scene, our framework ranks first with an average precision of 97.8\% and an average recall of 98.6\%, which demonstrates the effectiveness of our framework. 
\end{abstract}




\maketitle

\section{Introduction}

Accurate object detection is a fundamental necessity for autonomous or assisted driving. Many previous works~\cite{girshick2014rich,redmon2016you,tran2015learning,zhu2017flow,wang2021temporal} have achieved good performance based on visual images or videos captured by RGB cameras. 
However, camera-based methods can easily fail in bad driving conditions, such as frogging weather, dimming night, and strong lighting. Compared with visual light, LiDAR can provide direct and robust distance measurement of the surrounding environment~\cite{yue2018lidar, wang2019pseudo}, but LiDAR scanners are so expensive that many autonomous car manufacturers prefer not to use them. Similar to LiDAR, millimeter-wave can function reliably and detect range accurately; and similar to RGB camera, radar sensors are fairly competitive in terms of manufacturing cost. Therefore, object detection based on a frequency modulated continuous wave (FMCW) radar has been considered as a more robust and practical choice.

\begin{figure*}[t]
 \centering
 \includegraphics[width=6in]{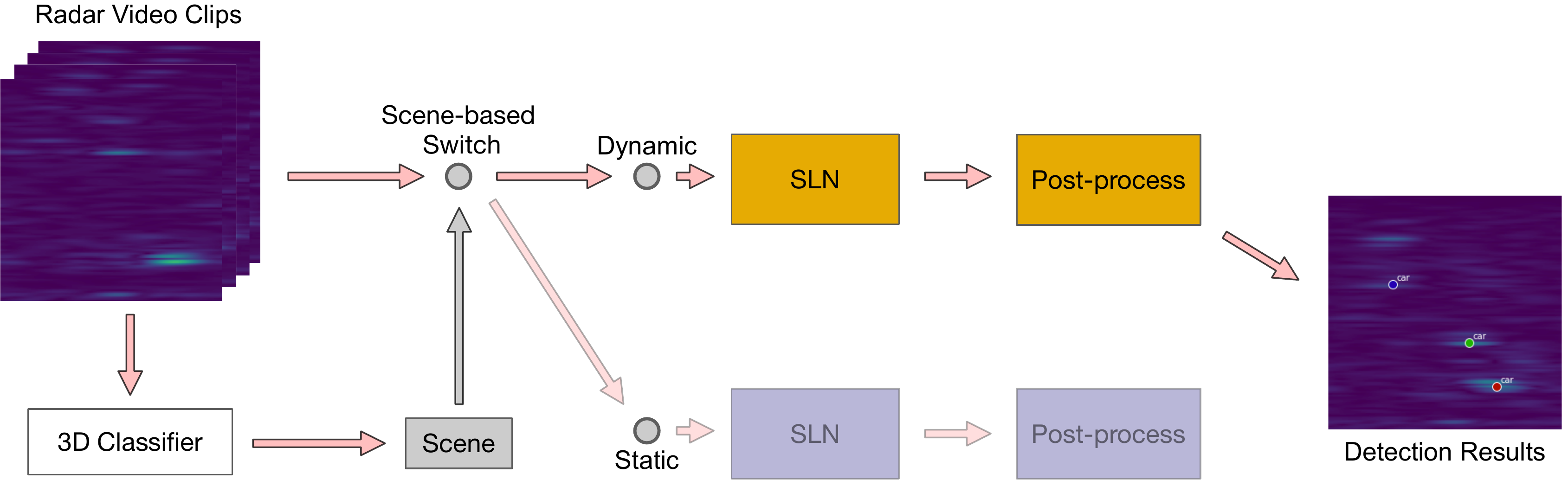}
 \caption{An overview of the scene-aware learning framework in test phase. Radar sequences are first classified into different scenes by a 3D-classifier. Then the scene-based switch will pass the radar snippet to the corresponding SLNet branch (\textcolor{dynamic}{orange}: Dynamic, \textcolor{static}{blue}: Static). SLNet trained on corresponding radar sequence will predict the ConfMaps of objects. Detection results are outputted with a scene-specific post-processing process.}
 \label{fig:test}
\end{figure*}

Compared with visual images, radar frequency (RF) data is much harder to annotate. In order for better representation, RF data are usually transformed into the format of range-azimuth frequency heatmaps (RAMaps), whose horizontal and vertical dimensions denote angle and distance (a bird-eye view), respectively. 
Recently, \cite{wang2021rodnet} proposed a pipeline for radar object detection, a cross-modal supervision framework that generates labels for RF data without laborious and inconsistent human labeling, which enables neural network training on a large amount of consistently annotated RF data. High-performance detection models are utilized for labeling on the RGB images and transform object positions into points on RAMaps. To train the models, annotations of objects are transformed into object confidence distribution maps (ConfMaps). During the test phase, the output ConfMaps will be processed to generate the detection results. To evaluate the final results, \cite{wang2021rodnet} defines an average precision metric similar to the one used in traditional object detection. 
Our framework follows a similar annotation generation process.

To perform detection on the RF data, \cite{wang2021rodnet} directly applies 3D-version of previous models \cite{szegedy2015going,newell2016stacked} for object detection without considering the inherent property of radar sequences. For example, more attention should be paid to the velocity information which can be retrieved from the RF data. The unique properties of RF data can provide us more understanding of the semantic meanings of an object. 


In this paper, we propose a branched scene-aware learning framework for radar object detection. Specifically, the framework consists of two parts: a scene classifier and a radar object detector. We find that RF data in different driving scenes exhibit significant differences. Therefore, we partition all radar sequences into different sets based on the driving scene. The scene classifier will predict the scene category for each input radar sequence, \textit{e.g.} static or moving background. The object detection branches are trained for two stages. In the first stage, a Scene-aware Learning Network (SLNet)
is first trained on all the RF sequences to learn a universal well-behaved object detector. In the second stage, for each type of scene, a scene-specific radar object detector is fine-tuned with the corresponding radar sequences on top of the universal model. As a result, the fine-tuned models are able to learn more scene-specific features for better performance.

Based on well-performed neural network architectures in video recognition, \textit{e.g.} Conv(2+1)D~\cite{tran2018closer} and ResNet~\cite{he2016deep}, we build different variants of SLNet. 
For better model generalization accuracy, we design and apply scene-aware augmentations SceneMix on the RF data during training. SceneMix creates a new training radar snippet by inserting a 
piece of one radar snippet into another snippet of the same scene category. More specifically, the snippet of the radar sequence can be mixed up, cropped and replaced, or de-noised and added with other radar snippets. To make our results more robust, we further design a new type of post-processing to the detection results and vary the process in different scenes.

We train and evaluate our network in ROD2021 challenge. The ROD2021 dataset in this challenge contains 40 sequences for training and 10 sequences for testing. We simply define two scenes: Static and Dynamic, which depend on whether the car carrying the radar sensor is moving or not. In this challenge, our SLNet can achieve about 75.0\% average precision (AP) and 81.0\% average recall (AR). Moreover, we achieve 97.8\% AP and 98.6\% AR in the Parking Lot category, \textbf{ranking first in the challenge}.

\begin{itemize}
\item We propose a novel scene-aware learning framework for radar object detection based on the type of driving scene.

\item We propose to leverage the spatio-temporal convolutional block "R(2+1)D" and build the Scene-aware Learning Network (SLNet) for accurate radar object detection.

\item We customize some image-processing methods for radar. Specifically, we propose novel augmentation, post-processing, and ensembling schemes for the new data modality. 
\end{itemize}

\section{Related Work}

\subsection{Object Detection for Images and Videos}
Convolution neural network has achieved remarkable performance in various computer vision tasks, including object detection for visual images and videos. Most state-of-the-art image-based object detection methods can be classified into two categories: multiple-stage pipeline and single-stage pipeline.

The classic model of multiple-stage is R-CNN~\cite{girshick2014rich}. In this setting, regions of interests (RoIs) are first generated by neural networks. Then in each RoI, detection results will be obtained from the corresponding features. Variants of this model \cite{girshick2015fast,ren2015faster,he2017mask} further improve the speed and accuracy of R-CNN. The single-stage pipelines, on the other hand, directly predict the results by a single convectional network, \textit{e.g.} YOLO \cite{redmon2016you}.

To capture the relationship between frames in videos, many works have proposed different methods. \cite{zhu2017flow,wang2018fully} introduce flow-based methods, which combine the flow information and features extracted on one frame to obtain the prediction. Wang \etal~\cite{wang2021temporal} propose memory and self-attention to extract information in the temporal dimension. A more direct way to learn spatiotemporal features is proposed in~\cite{tran2015learning}, which builds 3D convolutional networks to extract the features from radar snippets.

\begin{figure}[t]
 \centering
 \includegraphics[width=3.2in]{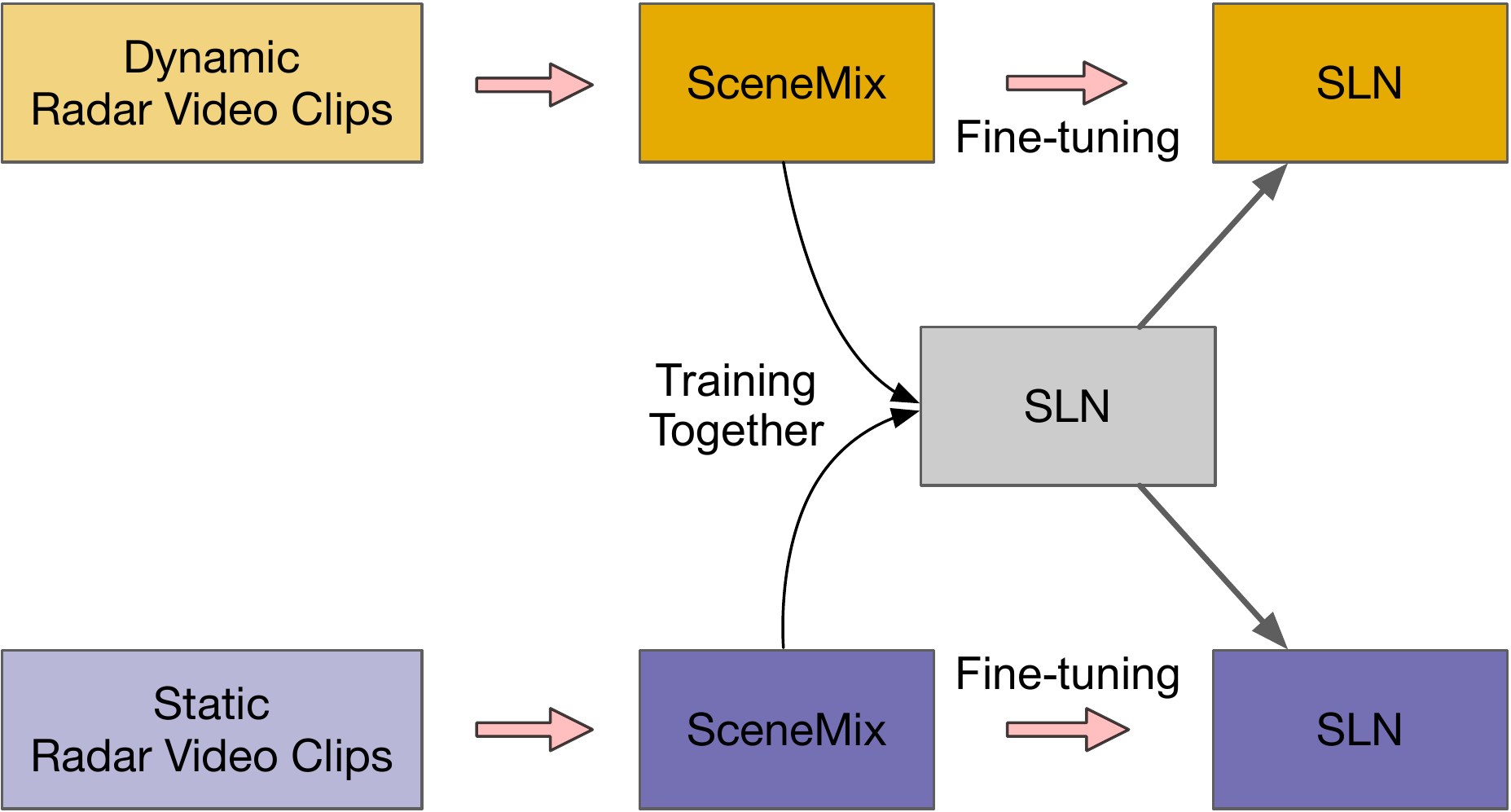}
 \caption{An overview of the scene-aware learning framework in training stage. Coloring represents different scenes (\textcolor{dynamic}{orange}: Dynamic, \textcolor{static}{blue}: Static).}
 \label{fig:train}
\end{figure}

\subsection{Radar Object Detection}

To overcome the bad quality of camera sensors in severe weather or unsatisfied lighting, some prior works~\cite{olver1988fmcw,gao2019experiments,nabati2019rrpn,nobis2019deep,danzer20192d,pham2018buried,manjunath2018radar} exploit the data from a Frequency Modulated Continuous Wave (FMCW) radar to detect the object more robustly. Considering it is hard to annotate the radar data since a human has less knowledge about what an object like in radar images, previous radar object detection works can be classified into two categories by whether visual data is required during learning.

The first type of radar object detection is to fuse radar and vision images together to obtain a more robust detection result. Nabati, \etal~\cite{nabati2019rrpn} fuses the data collected from radars with
vision data to obtain faster and more accurate detections. Meanwhile, Nobis \etal~\cite{nobis2019deep} extract and combine features of visual images and sparse radar data in the network encoding layers to improve the 2D object detection results. 

The second type is detecting objects based on radar data only. To effectively and efficiently collect radar object annotation, with a calibrated camera or LiDAR sensor, some annotations are automatically generated by high-accuracy object detection algorithms on these data~\cite{wang2021rodnet,meyer2019automotive}. 
Wang \etal~\cite{wang2021rodnet} proposes a cross-modal supervision pipeline to annotation radar sequences with less human labor and represent the radar frequency data in the rage-azimuth coordinates (RAMaps). This pipeline facilitates the development of a radar object detection algorithm.

\begin{figure}[t]
 \centering
 \includegraphics[width=3.2in]{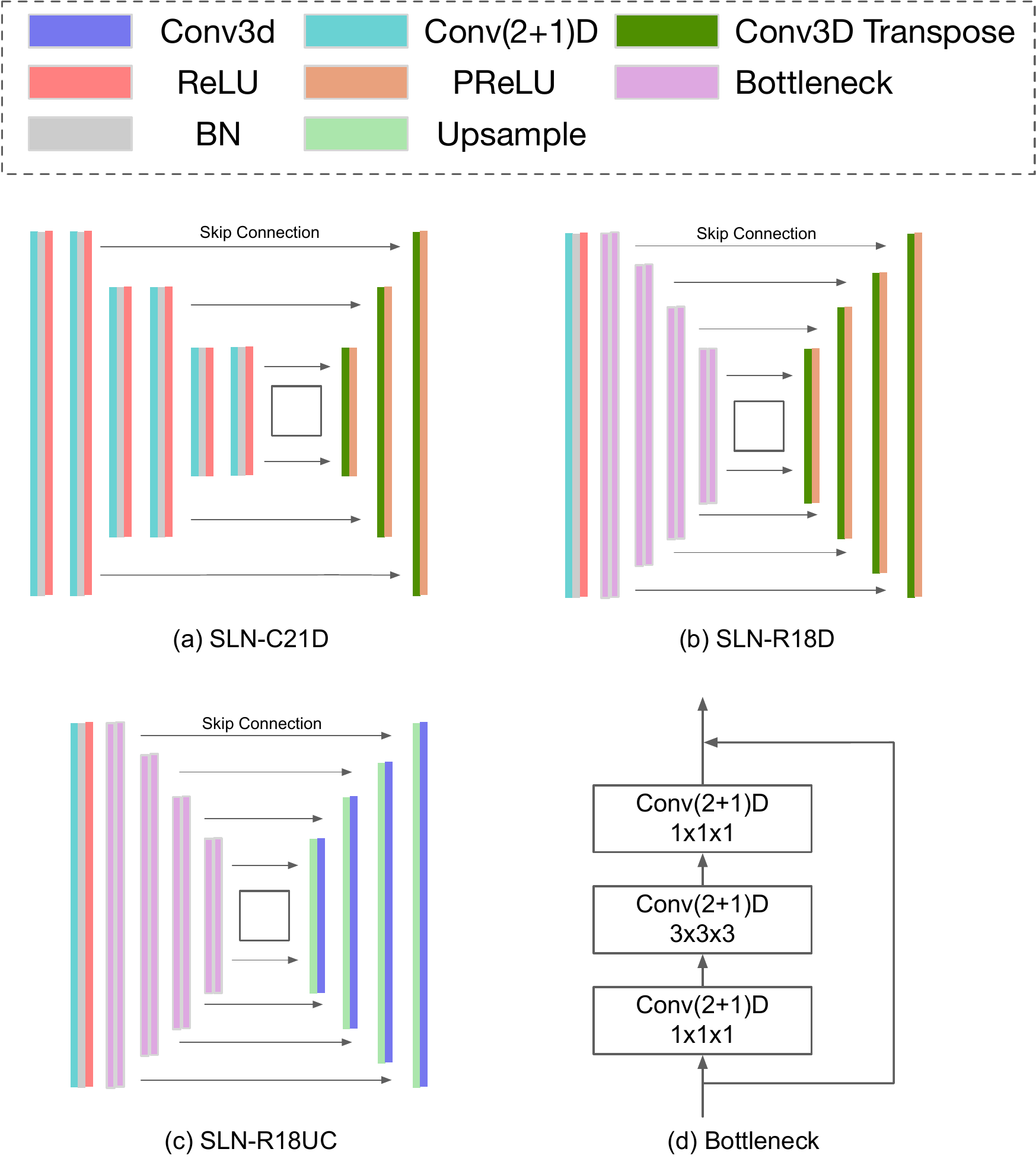}
 \caption{The architectures of our three SLNet models.}
 \label{fig:arch}
\end{figure}

\subsection{CNN for Radar Processing}
In the processing of radar data, a series of research \cite{angelov2018practical,capobianco2017vehicle,kwon2017human,qiao2020human,wang2021rodnet} explores convolution neural networks to extract features of radar data. To obtain good feature representations for radar data, Capobianco \etal~\cite{capobianco2017vehicle} apply a convolutional neural network to the rage Doppler signature. While Angelov \etal~\cite{angelov2018practical} try out various network structures, including residual networks and a combination of the convolutional and recurrent networks to classify radar objects. To prevent overfitting, Kwon \etal~\cite{kwon2017human} adds Gaussian noise to the input radar data. Since radar data are usually represented in the format of complex numbers,~\cite{gao2018enhanced} proposed to utilize complex-valued CNN to enhance radar recognition.

To better extract spatiotemporal information for radar object detection, prior works utilize 3D convolution on radar data. Hazara \etal~\cite{hazra2019short} propose to use 3D CNN architecture to learn the embedding model with a distance-based triplet-loss similarity metric. In~\cite{wang2021rodnet}, three encoder-decoder-based convolution network structures are proposed for radar object detection. The encoder consists of a series of 3D convolution layers and the decoder is composed of several transpose convolution layers.

\begin{figure*}[t]
 \centering
 \includegraphics[width=6in]{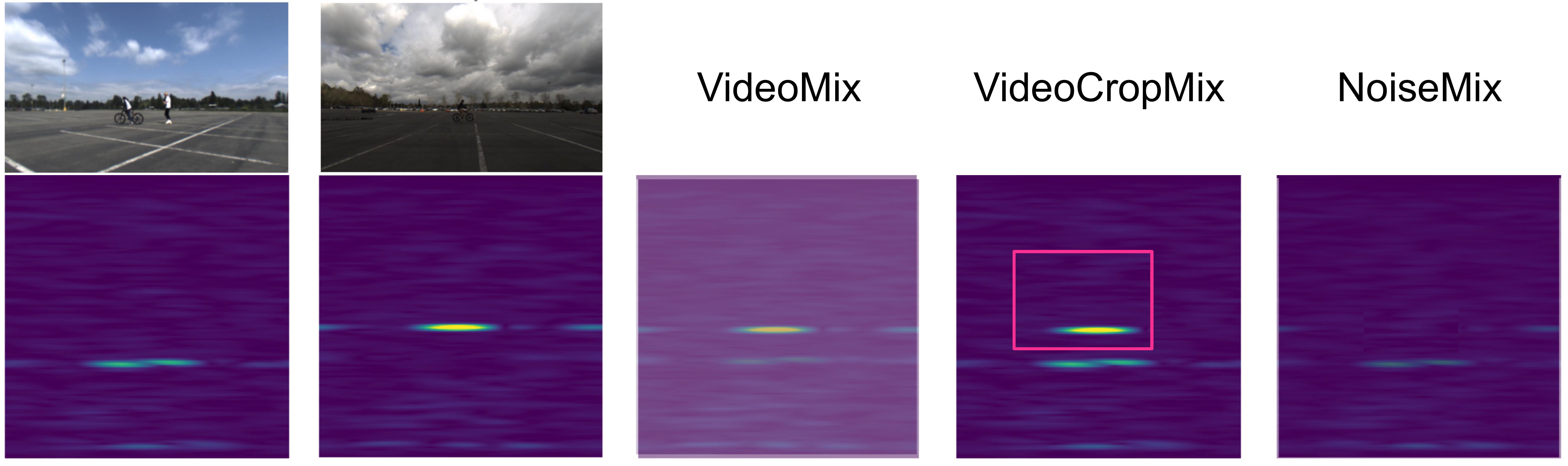}
 \caption{Example results of the SceneMix augmentation. The left two frames are of static scenes. The right three frames are the results of VideoMix, VideoCropMix and NoiseMix respectively.}
 \label{fig:aug}
\end{figure*}

\section{Approach}

Following~\cite{wang2021rodnet}, we formulate the radar objection detection as follows: with a training radar sequence in the format of RAMaps $R_{train}$ and its annotation (points with semantic class label) $y_{train}$, we are required to detect radar objects on the testing sequences $R_{test}$. ConfMap $C$ is generated from annotation $y$ for neural network supervision by utilizing Gaussian distributions to set the values around the object location. With a sliding window $\tau$, the network is fed with a snippet of $R_{train}$ with dimension $(C_{RF},\tau,\omega,h)$ and predicts a ConfMap $\hat C$ with dimension $(C_{cls},\tau,\omega,h)$. $C_{RF}$ is the number of channels in RAMaps, which consists of real and imaginary~\cite{zhao2018through}, while $C_{cls}$ is the number of object classes. In the test phase, the output ConfMaps are processed to point detect results $\hat y$. The performance of the model will be evaluated between $y_{test}$ and $\hat y$.

In this section, we will first introduce the scene-aware learning framework for radar object detection. Then the components in the framework, which are the architecture of Scene-aware Learning Network (SLNet), SceneMix, and Post-Processing, will be described in detail.

\subsection{Scene-aware Learning Framework}

Radar frequency data under different driving scenes differ greatly. One reason for this is the inherent velocity information in radar signals. Therefore, whether the ego car is moving or not will lead to a great difference in the signals of objects and noise. For example, the relative velocity of a car on the highway may be zero, while it can be very high if the ego car is static. Another is the different possibility of objects appearing in different scenes. Thus, we design a scene-aware learning framework to tackle this problem.

Specifically, we divide all RF data sequences into two scene categories: Dynamic and Static, depending on whether the ego car is moving or not. We adopt a two-stage training approach for the SLNet.
In the first phase, all radar snippets are used to train a universal SLNet (described in Section~\ref{arch}). In the second phase, we create two branches, each responsible for the radar object detection in one scene. In each branch, we fine-tune the SLNet based on the universal model obtained in the first phase with radar snippets of the corresponding scene.
The whole framework is shown in Figure~\ref{fig:train}.

We also train a scene classifier to classify the input radar snippets into one of these two scenes. With the classifier, the overall framework for scene-aware learning is shown in Figure~\ref{fig:test}. During the test phase, the scene classifier will first classify a test radar snippet into one of the two scene categories. 
Based on the scene category, the test radar snippet will then be fed into the corresponding SLNet branch to generate the ConfMap of radar objects. 
The scene-specific post-processing will be applied to the output ConfMaps of the SLNet (described in Section~\ref{post}) to get final detection results.

\begin{figure}[t]
 \centering
 \includegraphics[width=3.2in]{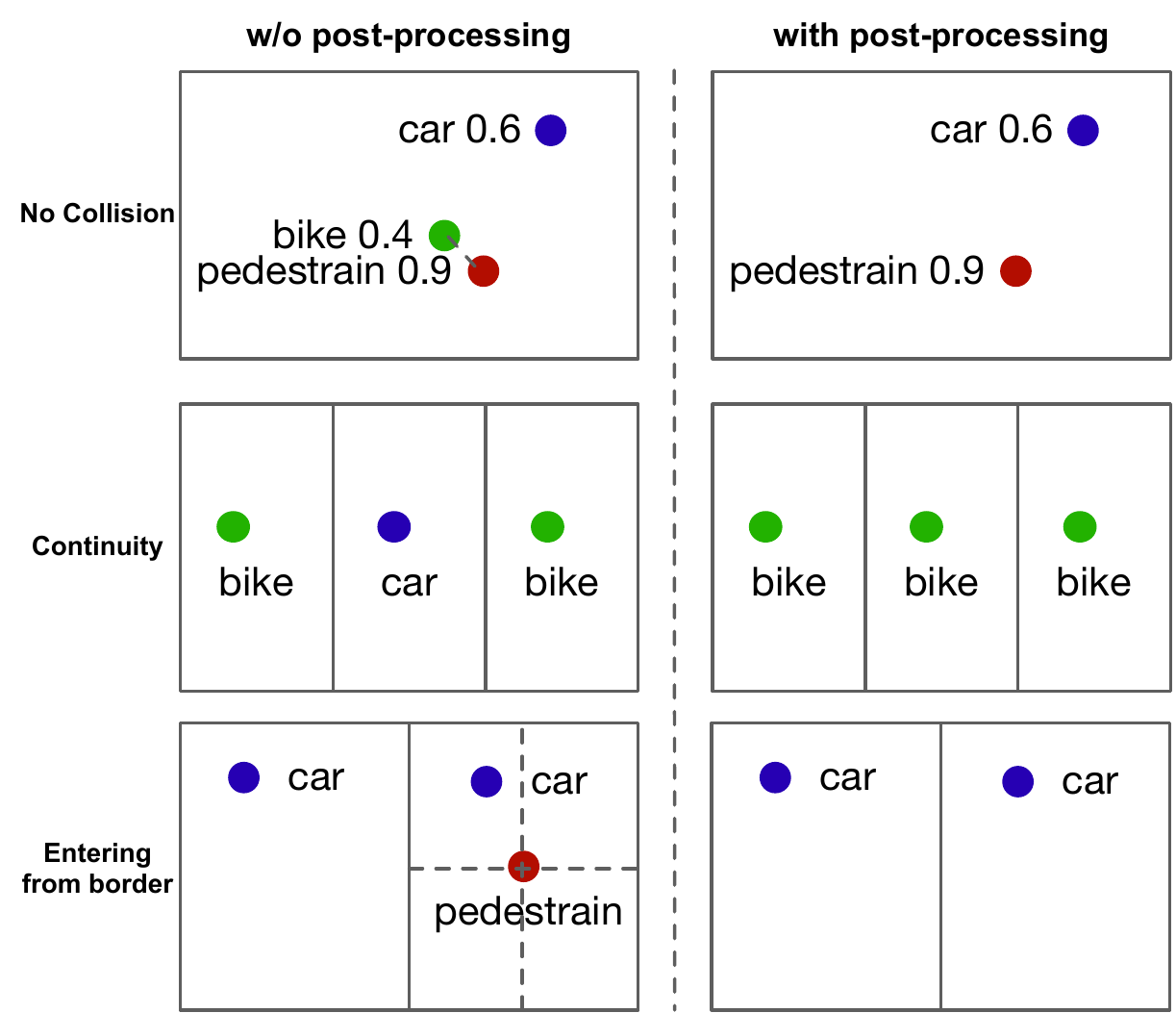}
 \caption{The examples of post-processing. An bounding box represents for one frame (different size of bounding boxes represents frames of the same size for better visualization). A sequence of frames are placed from left to right in time sequence. Different color of points represent different detection results. The number next to the class name stands for the confidence of the predicting results.}
 \label{fig:post}
\end{figure}

\begin{table*}[t]
\caption{Radar object detection performance on ROD2021 dataset.}
\begin{tabular}{@{}l|llllll|llllll@{}}
\toprule
Architectures & $\text{AP}$ & $\text{AP}^{0.5}$ & $\text{AP}^{0.6}$ & $\text{AP}^{0.7}$ & $\text{AP}^{0.8}$ & $\text{AP}^{0.9}$ & $\text{AR}$ & $\text{AR}^{0.5}$ & $\text{AR}^{0.6}$ & $\text{AR}^{0.7}$ & $\text{AR}^{0.8}$ & $\text{AR}^{0.9}$ \\ \midrule
RODNet-CDC & 45.38 & 50.89 & 49.62 & 47.81 & 43.85 & 31.69 & 50.74 & 54.90 & 53.83 & 52.58 & 49.39 & 40.89 \\
RODNet-HG & 41.28 & 47.66 & 46.49 & 44.63 & 39.64 & 25.01 & 47.83 & 52.74 & 51.80 & 50.23 & 46.57 & 35.50 \\
RODNet-HGwI & 38.82 & 44.26 & 42.28 & 40.63 & 37.44 & 27.22 & 45.96 & 50.23 & 48.66 & 47.17 & 44.79 & 37.44 \\ \midrule
SLNet-C21D & 46.84 & 52.32 & 51.02 & 49.36 & 45.33 & 33.12 & 52.23 & 56.45 & 55.29 & 53.96 & 50.91 & 42.59 \\
SLNet-R18D & 47.22 & 53.50 & 52.16 & 49.63 & 44.99 & 33.30 & 54.49 & 59.39 & 58.33 & 56.45 & 52.57 & 43.67 \\
SLNet-R18UC & 53.41 & 60.00 & 58.51 & 56.18 & 51.06 & 37.59 & 59.52 & 63.84 & 62.72 & 61.33 & 58.18 & 48.73 \\ \midrule
Ensemble & \textbf{54.15} & \textbf{59.89} & \textbf{58.96} & \textbf{56.53} & \textbf{52.10} & \textbf{40.39} & \textbf{60.84} & \textbf{65.37} & \textbf{64.28} & \textbf{62.57} & \textbf{59.09} & \textbf{50.76} \\ \bottomrule
\end{tabular}
\label{tab:rod2021}
\end{table*}

\begin{table}[]
\caption{Teams with high ranking and corresponding model performances in ROD2021 Challenge}
\begin{tabular}{@{}lccc@{}}
\toprule
Team & AP (total) & AR (total) & AP (PL) \\ \midrule
\multicolumn{1}{l}{Baidu-VIS\&ITD} & \textbf{82.2} & 90.1 & 97.0 \\
\multicolumn{1}{l}{USTC-NELSLIP} & 79.7 & \textbf{88.9} & 95.6 \\
\multicolumn{1}{l}{No\_Bug} & 76.1 & 83.9 & 96.1 \\
\multicolumn{1}{l}{DD\_Vision} & 75.1 & 84.9 & 95.2 \\
\multicolumn{1}{l}{\Gc Ours} & \Gc 75.0 & \Gc 81.0 & \multicolumn{1}{c}{\Gc \textbf{97.8}} \\
\multicolumn{1}{l}{acvlab} & 69.3 & 77.3 & 69.3 \\ \bottomrule
\end{tabular}
\label{tab:competition}
\end{table}

\subsection{Network Architecture}\label{arch}

We build three different network architectures for the ROD2021. The architectures are shown in Figure~\ref{fig:arch} with (2+1)D Convolution Deconvolution (SLNet-C21DC), ResNet(2+1)D18 Deconvolution (SLNet-R18D), and ResNet(2+1)D18 Upsamle-Convolution (SLNet-R18UC), respectively.

SLNet-CDC21 is adopted from RODNet-CDC~\cite{wang2021rodnet}, but we replace the 3D convolution in RODNet-CDC with (2+1)D convolution~\cite{tran2018closer}, and add shortcut connections according to~\cite{newell2016stacked}. Specifically, a (2+1)D convolutional block splits 3D convolution into a spatial 2D convolution followed by a temporal 1D convolution. Compared with 3D convolutional layer, a (2+1)D convolutional block introduces additional nonlinear rectification between
temporal and spatial convolution. Besides, the decomposition of temporal-spatial convolution facilitates the optimization according to~\cite{tran2018closer}.

SLNet-R18D substitutes the encoder of SLNet-C21DC with \\ResNet(2+1)D18~\cite{he2016deep}. We also utilize ResNet(2+1)D18 as the classifier to discriminate different scenes and this backbone is experimentally strong enough for this classification task.
As for SLNet-R18UC, the encoder is the same as SLNet-R18D while we adopt the structure of the decoder in~\cite{bulat2020toward}. The decoder is composed of upsampling and convolution instead of transposed convolution.

With sliced RAMap frames and ConfMaps, we train our SLNet with mean squared error loss:
\begin{equation}
    \mathcal{L}_{MSE}=-\sum_{cls}\sum_{i,j}(\hat C_{i,j}^{cls}-C_{i,j}^{cls})^2,
\end{equation}
where $C$ represents the ConfMaps generated from annotations, $\hat C$ represents the network prediction, and $C_{i,j}^{cls}$ represents the probability that object of class $cls$ appear at pixel $(i,j)$. 

Finally, we use an ensemble method on the aforementioned models to get the final results. Specifically, we average the ConfMaps of each model and then identify detection results from the averaged ConfMaps.

\subsection{SceneMix}\label{scenemix}

Many data augmentation methods have proven effective in different 2D and 3D tasks. VideoMix~\cite{yun2020videomix} and CutMix~\cite{yun2019cutmix} are powerful augmentation strategies that can both create new training snippets from two existing ones. These augmented images can not only enlarge our training dataset but also enforce our model to be more robust to different scenes.

We propose an augmentation method called scene-aware radar data mixing (SceneMix), which composes of VideoMix, VideoCropMix, and NoiseMix. Note that mixing radar snippets of different scenes may lead to absurd results, such as a static pedestrian in the Static scene will be running at the speed of a car on the highway if mixed to a Dynamic scene. Hence, only radar snippets of the same scene will be mixed together.

Denote radar snippet with $x\in\mathbb{R}^{C_{RF}\times T\times W\times H}$ and corresponding ConfMaps with $c\in\mathbb{R}^{C_{cls}\times T\times W\times H}$. The VideoMix algorithm mix two radar snippets with random proportion $\lambda\in [0,1]$. The new radar snippet is generated by:
\begin{equation}
    \begin{aligned}
        x &= \lambda x_A +(1-\lambda) x_B \\
        c &= \lambda c_A +(1-\lambda) c_B
    \end{aligned}
\end{equation}

The VideoCropMix algorithm mixes two radar snippets in another way: randomly crop on a radar snippet and replace the cropped area with the corresponding area in another video. The same process is also performed on the ConfMaps.

Adding noise to training samples has proven to help train a more robust neural network. To generate diverse radar noise, we introduce the NoiseMix augmentation. Notice that each radar snippet contains noisy signals naturally. To extract the noise from radar snippets, we set the area in which one of the semantic classes has a probability greater than a threshold in ConfMaps to zero. Then, the extracted noise is added to other radar snippets without modifying its ConfMaps.

\subsection{Post-Processing} \label{post}

After predicting ConfMaps from SLNet, post-processing needs to be applied to transform the ConfMaps into final detections. The L-NMS~\cite{wang2021rodnet} in the proposed pipeline is a good choice but fails to take the property of driving scenes into consideration. Apart from using L-NMS to identify detection from ConfMaps, we introduce a series of constraints to make the results more robust, including No Collision, Continuity, and Entering from the border. An illustration of the following post-processing constraints is shown in Figure~\ref{fig:post}.

{\bf No Collision}: If two objects of the different classes are close to each other, then the less confident one is removed to prevent a collision. We measure the distance of two objects by object location similarity (OLS)~\cite{wang2021rodnet}.

{\bf Continuity}: If one object appears continuously in frames but gets missing or changes into another class in one or two frames among them, then we use linear interpolation to add or change the class of the object to those frames.

{\bf Entering from the border}: If one object appears suddenly (which means cannot be tracked back in frames to the border of the radar image), then we consider it noise and delete it.

All three constraints are applied to the outputs in Static scenes. For Dynamic scenes, we find the last two constrain have little effect due to the fast speed of the vehicle carrying the radar sensor. Thus, we only apply the first constraint in this scene.

\begin{table*}[t]
\caption{Ablation study on different components of the framework. The vallina version of SLNet-R18UC is trained in a direct way. $\text{AP}^{S}$ and $\text{AR}^{S}$ means the AP and AR on sequences of Static scenes while $\text{AP}^{D}$ and $\text{AR}^{D}$ denote those of Dynamic ones.}
\begin{tabular}{@{}lllllll@{}}
\toprule
Methods & $\text{AP}$ & $\text{AP}^{S}$ & $\text{AP}^{D}$ & $\text{AR}$ & $\text{AR}^{S}$ & $\text{AR}^{D}$ \\ \midrule
SLNet-R18UC (vallina) & 47.03 & 70.51 & 18.47 & 46.95 & 75.07 & 27.40 \\
with SceneMix & 49.97 & 73.55 & 23.16 & 55.94 & 78.09 & 32.14 \\
with Fine-tuning on S & 52.69 & 74.23 & 22.91 & 58.85 & 78.00 & 30.98 \\
with Fine-tuning on D & 50.49 & 73.32 & \textbf{28.27}  & 56.09 & 77.51 & \textbf{37.85}  \\ \midrule
SLNet-R18UC & \textbf{53.41} & \textbf{74.23} & \textbf{28.27} & \textbf{60.00} & \textbf{78.00} & \textbf{37.85} \\ \bottomrule
\end{tabular}
\label{tab:ablation}
\end{table*}

\section{Experiments}

\subsection{Datasets}

The \textbf{ROD2021} dataset used in ROD2021 Challenge is a subset of \textbf{CRUW}~\cite{wang2021rodnet}
dataset. There are 50 sequences in total, where 40 of them are provided with annotations. Each sequence lasts around 25-60s with 800-1700 frames. Each frame is a RAMap with a dimension 128$\times$128. The provided annotation is created by a camera-radar fusion algorithm \cite{wang2021rodnet}. 

To validate our algorithm, we randomly choose 8 sequences from 40 sequences with annotations as the validation set and the rest 32 sequences as the training set. Among 40 sequences, about 15\% are classified as Dynamic and the rest are Static. In additional to presenting the performance on test set from ROD2021 competition server, we also provide more detailed analysis by conducting experiments on the validation set.

\subsection{Evaluation Metrics}

To evaluate our methods, we use the average precision (AP) and average recall (AR) metrics proposed in \cite{wang2021rodnet}. Specifically, the object location similarity (OLS) \cite{wang2021rodnet} between our detection results and ground truth are calculated. Then, with threshold $t$, detection results with OLS higher than $t$ are considered a correct match and thus the precision and recall can be computed. With $t$ ranging from 0.5 to 0.9 with a step of 0.05, we get the AP and AR as our evaluation metrics.


\subsection{Training Details}

Our experiments utilize Adam~\cite{kingma2014adam} to optimize the network, and the learning rate is set to $1\times10^{-4}$. A cosine annealing with warmup restart scheduler is applied on the optimizer in order to make the training process more smooth. The model is paralleled on 4 GPUs with batch size 64 in total. Given an input radar snippet, the probability of applying VideoMix and VideoCropMix are both $\frac{1}{3}$. The chance of augmentation with NoiseMix is $\frac{1}{2}$. After 50 epochs of training on all sequences, our model is then fine-tuned on sequences of different scenes for 30 epochs. Finally, all trained models are ensembled to get the detection results.

\begin{table}[t]
\caption{Number of different scenes and prediction accuracy.}
\begin{tabular}{@{}lll@{}}
\toprule
Scene & Static & Dynamic \\ \midrule
\# seq in train & 28 & 4 \\
\# seq in test & 6 & 2 \\
Accuracy & 100\% & 100\% \\ \bottomrule
\end{tabular}
\label{tab:scene}
\end{table}

\subsection{Results}

Table~\ref{tab:competition} presents final results of models with high ranking in ROD2021 competition. Our model achieves a AP of 75.0\%, which outperforms a baseline of 69.8\% by simply applying RODNet-CDC without the scene-aware learning framework. It is worth noting that our model ranked first in Parking Lot(PL) scene with respect to the AP score.

To further compare performance of different models, we utilize the validation set to compare our methods with the three models proposed in \cite{wang2021rodnet}. The results are shown in Table~\ref{tab:rod2021}. RODNet-CDC is a shallow 3D CNN encoder-decoder network. RODNet-HG is adopted from~\cite{newell2016stacked} with only one stack, while RODNet-HGwI replaces the 3D convolution layers in RODNet-HG with temporal inception layers \cite{szegedy2015going}.

To complete the scene-aware learning pipeline, we need to train a 3D scene classifier. The number of different scenes in train and test datasets are shown in Table~\ref{tab:scene}. Our scene classifier can obtain 100\% accuracy in predicting the driving scenes.

All results of SLNet in Table~\ref{tab:rod2021} are trained in the scene-aware learning framework. We can see that the ensemble version of the scene-aware learning framework outperforms the best results of baselines 8.77\% in average precision and 10.10\% in average recall. All SLNet (RODNet-CDC, SLNet-R18C, and SLNet-R18UD) outperforms three baselines by 1.46\%, 1.84\%, and 8.03\% in AP respectively.


\subsection{Ablation Study}

Next, we investigate the effectiveness of different components in the scene-aware learning framework on the validation set of ROD2021 dataset. Table \ref{tab:ablation} shows the results with and without SceneMix augmentation, and also the results fine-tuning on different scenes. 

The vallina version of SLNet-R18UC is trained directly without fine-tuning and SceneMix augmentations. When training with SceneMix, the final AP increases by 1.77\%. Based on the SLNet-R18UC trained with SceneMix, we fine-tune on Static and Dynamic scenes separately, and both of the fine-tuned models achieve a better result by 2.72\% and 0.52\% in AP respectively. Besides, fine-tuning on different scenes will lead to an obvious improvement in the corresponding scene.

Finally, by applying the scene-aware learning framework, we predict each scene with the corresponding model and achieve final AP of 53.41\%. We can observe that adding each component contributes to the final results without any performance degradation.

\section{Discussion}
The scene-aware learning framework can remarkably improve the performance of radar object detection, especially for the Static scene. Despite the success of this model, there are also some limitations which need further attention. First, although scene-aware learning achieves high accuracy on Static scene, the performance on Dynamic ones is not satisfactory enough. More analysis should be done to investigate why this method has inferior performance in other scenarios like campus road, city street, and highways. Besides, apart from two scenes division, it may be possible for the model to generalize to more categories, or even velocity-aware one. Finally, how to apply this model to an unseen scene may be another practical issue. We leave these questions for future work.

\section{Conclusion}
In this paper, we proposed a scene-aware learning framework to detect objects from radar sequences. In the framework, radar sequences will be detected by models fine-tuned on the same scenes. The proposed SLNet can robustly detect objects with high precision. In addition, the paper presents a new augmentation SceneMix and post-processing method for radar object detection. The proposed method offers a novel and effective solution to take advantage of the properties of radar data. Our experiments conducted on the ROD2021 dataset demonstrate our proposed framework is an accurate and robust method to detect objects based on radar.


\bibliographystyle{ACM-Reference-Format}
\balance
\bibliography{icmr}


\begin{thebibliography}{35}


\ifx \showCODEN    \undefined \def \showCODEN     #1{\unskip}     \fi
\ifx \showDOI      \undefined \def \showDOI       #1{#1}\fi
\ifx \showISBNx    \undefined \def \showISBNx     #1{\unskip}     \fi
\ifx \showISBNxiii \undefined \def \showISBNxiii  #1{\unskip}     \fi
\ifx \showISSN     \undefined \def \showISSN      #1{\unskip}     \fi
\ifx \showLCCN     \undefined \def \showLCCN      #1{\unskip}     \fi
\ifx \shownote     \undefined \def \shownote      #1{#1}          \fi
\ifx \showarticletitle \undefined \def \showarticletitle #1{#1}   \fi
\ifx \showURL      \undefined \def \showURL       {\relax}        \fi
\providecommand\bibfield[2]{#2}
\providecommand\bibinfo[2]{#2}
\providecommand\natexlab[1]{#1}
\providecommand\showeprint[2][]{arXiv:#2}

\bibitem[\protect\citeauthoryear{Angelov, Robertson, Murray-Smith, and
  Fioranelli}{Angelov et~al\mbox{.}}{2018}]%
        {angelov2018practical}
\bibfield{author}{\bibinfo{person}{Aleksandar Angelov}, \bibinfo{person}{Andrew
  Robertson}, \bibinfo{person}{Roderick Murray-Smith}, {and}
  \bibinfo{person}{Francesco Fioranelli}.} \bibinfo{year}{2018}\natexlab{}.
\newblock \showarticletitle{Practical classification of different moving
  targets using automotive radar and deep neural networks}.
\newblock \bibinfo{journal}{\emph{IET Radar, Sonar \& Navigation}}
  \bibinfo{volume}{12}, \bibinfo{number}{10} (\bibinfo{year}{2018}),
  \bibinfo{pages}{1082--1089}.
\newblock


\bibitem[\protect\citeauthoryear{Bulat, Kossaifi, Tzimiropoulos, and
  Pantic}{Bulat et~al\mbox{.}}{2020}]%
        {bulat2020toward}
\bibfield{author}{\bibinfo{person}{Adrian Bulat}, \bibinfo{person}{Jean
  Kossaifi}, \bibinfo{person}{Georgios Tzimiropoulos}, {and}
  \bibinfo{person}{Maja Pantic}.} \bibinfo{year}{2020}\natexlab{}.
\newblock \showarticletitle{Toward fast and accurate human pose estimation via
  soft-gated skip connections}.
\newblock \bibinfo{journal}{\emph{arXiv preprint arXiv:2002.11098}}
  (\bibinfo{year}{2020}).
\newblock


\bibitem[\protect\citeauthoryear{Capobianco, Facheris, Cuccoli, and
  Marinai}{Capobianco et~al\mbox{.}}{2017}]%
        {capobianco2017vehicle}
\bibfield{author}{\bibinfo{person}{Samuele Capobianco}, \bibinfo{person}{Luca
  Facheris}, \bibinfo{person}{Fabrizio Cuccoli}, {and} \bibinfo{person}{Simone
  Marinai}.} \bibinfo{year}{2017}\natexlab{}.
\newblock \showarticletitle{Vehicle classification based on convolutional
  networks applied to fmcw radar signals}. In \bibinfo{booktitle}{\emph{Italian
  Conference for the Traffic Police}}. Springer, \bibinfo{pages}{115--128}.
\newblock


\bibitem[\protect\citeauthoryear{Danzer, Griebel, Bach, and Dietmayer}{Danzer
  et~al\mbox{.}}{2019}]%
        {danzer20192d}
\bibfield{author}{\bibinfo{person}{Andreas Danzer}, \bibinfo{person}{Thomas
  Griebel}, \bibinfo{person}{Martin Bach}, {and} \bibinfo{person}{Klaus
  Dietmayer}.} \bibinfo{year}{2019}\natexlab{}.
\newblock \showarticletitle{2d car detection in radar data with pointnets}. In
  \bibinfo{booktitle}{\emph{2019 IEEE Intelligent Transportation Systems
  Conference (ITSC)}}. IEEE, \bibinfo{pages}{61--66}.
\newblock


\bibitem[\protect\citeauthoryear{Gao, Deng, Qin, Wang, and Li}{Gao
  et~al\mbox{.}}{2018}]%
        {gao2018enhanced}
\bibfield{author}{\bibinfo{person}{Jingkun Gao}, \bibinfo{person}{Bin Deng},
  \bibinfo{person}{Yuliang Qin}, \bibinfo{person}{Hongqiang Wang}, {and}
  \bibinfo{person}{Xiang Li}.} \bibinfo{year}{2018}\natexlab{}.
\newblock \showarticletitle{Enhanced radar imaging using a complex-valued
  convolutional neural network}.
\newblock \bibinfo{journal}{\emph{IEEE Geoscience and Remote Sensing Letters}}
  \bibinfo{volume}{16}, \bibinfo{number}{1} (\bibinfo{year}{2018}),
  \bibinfo{pages}{35--39}.
\newblock


\bibitem[\protect\citeauthoryear{Gao, Xing, Roy, and Liu}{Gao
  et~al\mbox{.}}{2019}]%
        {gao2019experiments}
\bibfield{author}{\bibinfo{person}{Xiangyu Gao}, \bibinfo{person}{Guanbin
  Xing}, \bibinfo{person}{Sumit Roy}, {and} \bibinfo{person}{Hui Liu}.}
  \bibinfo{year}{2019}\natexlab{}.
\newblock \showarticletitle{Experiments with mmwave automotive radar test-bed}.
  In \bibinfo{booktitle}{\emph{2019 53rd Asilomar Conference on Signals,
  Systems, and Computers}}. IEEE, \bibinfo{pages}{1--6}.
\newblock


\bibitem[\protect\citeauthoryear{Girshick}{Girshick}{2015}]%
        {girshick2015fast}
\bibfield{author}{\bibinfo{person}{Ross Girshick}.}
  \bibinfo{year}{2015}\natexlab{}.
\newblock \showarticletitle{Fast r-cnn}. In
  \bibinfo{booktitle}{\emph{Proceedings of the IEEE international conference on
  computer vision}}. \bibinfo{pages}{1440--1448}.
\newblock


\bibitem[\protect\citeauthoryear{Girshick, Donahue, Darrell, and
  Malik}{Girshick et~al\mbox{.}}{2014}]%
        {girshick2014rich}
\bibfield{author}{\bibinfo{person}{Ross Girshick}, \bibinfo{person}{Jeff
  Donahue}, \bibinfo{person}{Trevor Darrell}, {and} \bibinfo{person}{Jitendra
  Malik}.} \bibinfo{year}{2014}\natexlab{}.
\newblock \showarticletitle{Rich feature hierarchies for accurate object
  detection and semantic segmentation}. In
  \bibinfo{booktitle}{\emph{Proceedings of the IEEE conference on computer
  vision and pattern recognition}}. \bibinfo{pages}{580--587}.
\newblock


\bibitem[\protect\citeauthoryear{Hazra and Santra}{Hazra and Santra}{2019}]%
        {hazra2019short}
\bibfield{author}{\bibinfo{person}{Souvik Hazra} {and} \bibinfo{person}{Avik
  Santra}.} \bibinfo{year}{2019}\natexlab{}.
\newblock \showarticletitle{Short-range radar-based gesture recognition system
  using 3D CNN with triplet loss}.
\newblock \bibinfo{journal}{\emph{IEEE Access}}  \bibinfo{volume}{7}
  (\bibinfo{year}{2019}), \bibinfo{pages}{125623--125633}.
\newblock


\bibitem[\protect\citeauthoryear{He, Gkioxari, Doll{\'a}r, and Girshick}{He
  et~al\mbox{.}}{2017}]%
        {he2017mask}
\bibfield{author}{\bibinfo{person}{Kaiming He}, \bibinfo{person}{Georgia
  Gkioxari}, \bibinfo{person}{Piotr Doll{\'a}r}, {and} \bibinfo{person}{Ross
  Girshick}.} \bibinfo{year}{2017}\natexlab{}.
\newblock \showarticletitle{Mask r-cnn}. In
  \bibinfo{booktitle}{\emph{Proceedings of the IEEE international conference on
  computer vision}}. \bibinfo{pages}{2961--2969}.
\newblock


\bibitem[\protect\citeauthoryear{He, Zhang, Ren, and Sun}{He
  et~al\mbox{.}}{2016}]%
        {he2016deep}
\bibfield{author}{\bibinfo{person}{Kaiming He}, \bibinfo{person}{Xiangyu
  Zhang}, \bibinfo{person}{Shaoqing Ren}, {and} \bibinfo{person}{Jian Sun}.}
  \bibinfo{year}{2016}\natexlab{}.
\newblock \showarticletitle{Deep residual learning for image recognition}. In
  \bibinfo{booktitle}{\emph{Proceedings of the IEEE conference on computer
  vision and pattern recognition}}. \bibinfo{pages}{770--778}.
\newblock


\bibitem[\protect\citeauthoryear{Kingma and Ba}{Kingma and Ba}{2014}]%
        {kingma2014adam}
\bibfield{author}{\bibinfo{person}{Diederik~P Kingma} {and}
  \bibinfo{person}{Jimmy Ba}.} \bibinfo{year}{2014}\natexlab{}.
\newblock \showarticletitle{Adam: A method for stochastic optimization}.
\newblock \bibinfo{journal}{\emph{arXiv preprint arXiv:1412.6980}}
  (\bibinfo{year}{2014}).
\newblock


\bibitem[\protect\citeauthoryear{Kwon and Kwak}{Kwon and Kwak}{2017}]%
        {kwon2017human}
\bibfield{author}{\bibinfo{person}{Jihoon Kwon} {and} \bibinfo{person}{Nojun
  Kwak}.} \bibinfo{year}{2017}\natexlab{}.
\newblock \showarticletitle{Human detection by neural networks using a low-cost
  short-range Doppler radar sensor}. In \bibinfo{booktitle}{\emph{2017 IEEE
  Radar Conference (RadarConf)}}. IEEE, \bibinfo{pages}{0755--0760}.
\newblock


\bibitem[\protect\citeauthoryear{Manjunath, Liu, Henriques, and
  Engstle}{Manjunath et~al\mbox{.}}{2018}]%
        {manjunath2018radar}
\bibfield{author}{\bibinfo{person}{Ankith Manjunath}, \bibinfo{person}{Ying
  Liu}, \bibinfo{person}{Bernardo Henriques}, {and} \bibinfo{person}{Armin
  Engstle}.} \bibinfo{year}{2018}\natexlab{}.
\newblock \showarticletitle{Radar based object detection and tracking for
  autonomous driving}. In \bibinfo{booktitle}{\emph{2018 IEEE MTT-S
  International Conference on Microwaves for Intelligent Mobility (ICMIM)}}.
  IEEE, \bibinfo{pages}{1--4}.
\newblock


\bibitem[\protect\citeauthoryear{Meyer and Kuschk}{Meyer and Kuschk}{2019}]%
        {meyer2019automotive}
\bibfield{author}{\bibinfo{person}{Michael Meyer} {and} \bibinfo{person}{Georg
  Kuschk}.} \bibinfo{year}{2019}\natexlab{}.
\newblock \showarticletitle{Automotive radar dataset for deep learning based 3d
  object detection}. In \bibinfo{booktitle}{\emph{2019 16th European Radar
  Conference (EuRAD)}}. IEEE, \bibinfo{pages}{129--132}.
\newblock


\bibitem[\protect\citeauthoryear{Nabati and Qi}{Nabati and Qi}{2019}]%
        {nabati2019rrpn}
\bibfield{author}{\bibinfo{person}{Ramin Nabati} {and} \bibinfo{person}{Hairong
  Qi}.} \bibinfo{year}{2019}\natexlab{}.
\newblock \showarticletitle{Rrpn: Radar region proposal network for object
  detection in autonomous vehicles}. In \bibinfo{booktitle}{\emph{2019 IEEE
  International Conference on Image Processing (ICIP)}}. IEEE,
  \bibinfo{pages}{3093--3097}.
\newblock


\bibitem[\protect\citeauthoryear{Newell, Yang, and Deng}{Newell
  et~al\mbox{.}}{2016}]%
        {newell2016stacked}
\bibfield{author}{\bibinfo{person}{Alejandro Newell}, \bibinfo{person}{Kaiyu
  Yang}, {and} \bibinfo{person}{Jia Deng}.} \bibinfo{year}{2016}\natexlab{}.
\newblock \showarticletitle{Stacked hourglass networks for human pose
  estimation}. In \bibinfo{booktitle}{\emph{European conference on computer
  vision}}. Springer, \bibinfo{pages}{483--499}.
\newblock


\bibitem[\protect\citeauthoryear{Nobis, Geisslinger, Weber, Betz, and
  Lienkamp}{Nobis et~al\mbox{.}}{2019}]%
        {nobis2019deep}
\bibfield{author}{\bibinfo{person}{Felix Nobis}, \bibinfo{person}{Maximilian
  Geisslinger}, \bibinfo{person}{Markus Weber}, \bibinfo{person}{Johannes
  Betz}, {and} \bibinfo{person}{Markus Lienkamp}.}
  \bibinfo{year}{2019}\natexlab{}.
\newblock \showarticletitle{A deep learning-based radar and camera sensor
  fusion architecture for object detection}. In \bibinfo{booktitle}{\emph{2019
  Sensor Data Fusion: Trends, Solutions, Applications (SDF)}}. IEEE,
  \bibinfo{pages}{1--7}.
\newblock


\bibitem[\protect\citeauthoryear{Olver and Cuthbert}{Olver and
  Cuthbert}{1988}]%
        {olver1988fmcw}
\bibfield{author}{\bibinfo{person}{AD Olver} {and} \bibinfo{person}{LG
  Cuthbert}.} \bibinfo{year}{1988}\natexlab{}.
\newblock \showarticletitle{FMCW radar for hidden object detection}. In
  \bibinfo{booktitle}{\emph{IEE Proceedings F (Communications, Radar and Signal
  Processing)}}, Vol.~\bibinfo{volume}{135}. IET, \bibinfo{pages}{354--361}.
\newblock


\bibitem[\protect\citeauthoryear{Pham and Lef{\`e}vre}{Pham and
  Lef{\`e}vre}{2018}]%
        {pham2018buried}
\bibfield{author}{\bibinfo{person}{Minh-Tan Pham} {and}
  \bibinfo{person}{S{\'e}bastien Lef{\`e}vre}.}
  \bibinfo{year}{2018}\natexlab{}.
\newblock \showarticletitle{Buried object detection from B-scan ground
  penetrating radar data using Faster-RCNN}. In
  \bibinfo{booktitle}{\emph{IGARSS 2018-2018 IEEE International Geoscience and
  Remote Sensing Symposium}}. IEEE, \bibinfo{pages}{6804--6807}.
\newblock


\bibitem[\protect\citeauthoryear{Qiao, Shan, and Tao}{Qiao
  et~al\mbox{.}}{2020}]%
        {qiao2020human}
\bibfield{author}{\bibinfo{person}{Xingshuai Qiao}, \bibinfo{person}{Tao Shan},
  {and} \bibinfo{person}{Ran Tao}.} \bibinfo{year}{2020}\natexlab{}.
\newblock \showarticletitle{Human identification based on radar micro-Doppler
  signatures separation}.
\newblock \bibinfo{journal}{\emph{Electronics Letters}} \bibinfo{volume}{56},
  \bibinfo{number}{4} (\bibinfo{year}{2020}), \bibinfo{pages}{195--196}.
\newblock


\bibitem[\protect\citeauthoryear{Redmon, Divvala, Girshick, and Farhadi}{Redmon
  et~al\mbox{.}}{2016}]%
        {redmon2016you}
\bibfield{author}{\bibinfo{person}{Joseph Redmon}, \bibinfo{person}{Santosh
  Divvala}, \bibinfo{person}{Ross Girshick}, {and} \bibinfo{person}{Ali
  Farhadi}.} \bibinfo{year}{2016}\natexlab{}.
\newblock \showarticletitle{You only look once: Unified, real-time object
  detection}. In \bibinfo{booktitle}{\emph{Proceedings of the IEEE conference
  on computer vision and pattern recognition}}. \bibinfo{pages}{779--788}.
\newblock


\bibitem[\protect\citeauthoryear{Ren, He, Girshick, and Sun}{Ren
  et~al\mbox{.}}{2015}]%
        {ren2015faster}
\bibfield{author}{\bibinfo{person}{Shaoqing Ren}, \bibinfo{person}{Kaiming He},
  \bibinfo{person}{Ross Girshick}, {and} \bibinfo{person}{Jian Sun}.}
  \bibinfo{year}{2015}\natexlab{}.
\newblock \showarticletitle{Faster r-cnn: Towards real-time object detection
  with region proposal networks}.
\newblock \bibinfo{journal}{\emph{arXiv preprint arXiv:1506.01497}}
  (\bibinfo{year}{2015}).
\newblock


\bibitem[\protect\citeauthoryear{Szegedy, Liu, Jia, Sermanet, Reed, Anguelov,
  Erhan, Vanhoucke, and Rabinovich}{Szegedy et~al\mbox{.}}{2015}]%
        {szegedy2015going}
\bibfield{author}{\bibinfo{person}{Christian Szegedy}, \bibinfo{person}{Wei
  Liu}, \bibinfo{person}{Yangqing Jia}, \bibinfo{person}{Pierre Sermanet},
  \bibinfo{person}{Scott Reed}, \bibinfo{person}{Dragomir Anguelov},
  \bibinfo{person}{Dumitru Erhan}, \bibinfo{person}{Vincent Vanhoucke}, {and}
  \bibinfo{person}{Andrew Rabinovich}.} \bibinfo{year}{2015}\natexlab{}.
\newblock \showarticletitle{Going deeper with convolutions}. In
  \bibinfo{booktitle}{\emph{Proceedings of the IEEE conference on computer
  vision and pattern recognition}}. \bibinfo{pages}{1--9}.
\newblock


\bibitem[\protect\citeauthoryear{Tran, Bourdev, Fergus, Torresani, and
  Paluri}{Tran et~al\mbox{.}}{2015}]%
        {tran2015learning}
\bibfield{author}{\bibinfo{person}{Du Tran}, \bibinfo{person}{Lubomir Bourdev},
  \bibinfo{person}{Rob Fergus}, \bibinfo{person}{Lorenzo Torresani}, {and}
  \bibinfo{person}{Manohar Paluri}.} \bibinfo{year}{2015}\natexlab{}.
\newblock \showarticletitle{Learning spatiotemporal features with 3d
  convolutional networks}. In \bibinfo{booktitle}{\emph{Proceedings of the IEEE
  international conference on computer vision}}. \bibinfo{pages}{4489--4497}.
\newblock


\bibitem[\protect\citeauthoryear{Tran, Wang, Torresani, Ray, LeCun, and
  Paluri}{Tran et~al\mbox{.}}{2018}]%
        {tran2018closer}
\bibfield{author}{\bibinfo{person}{Du Tran}, \bibinfo{person}{Heng Wang},
  \bibinfo{person}{Lorenzo Torresani}, \bibinfo{person}{Jamie Ray},
  \bibinfo{person}{Yann LeCun}, {and} \bibinfo{person}{Manohar Paluri}.}
  \bibinfo{year}{2018}\natexlab{}.
\newblock \showarticletitle{A closer look at spatiotemporal convolutions for
  action recognition}. In \bibinfo{booktitle}{\emph{CVPR}}.
  \bibinfo{pages}{6450--6459}.
\newblock


\bibitem[\protect\citeauthoryear{Wang, Wang, and Liu}{Wang
  et~al\mbox{.}}{2021b}]%
        {wang2021temporal}
\bibfield{author}{\bibinfo{person}{Hao Wang}, \bibinfo{person}{Weining Wang},
  {and} \bibinfo{person}{Jing Liu}.} \bibinfo{year}{2021}\natexlab{b}.
\newblock \showarticletitle{Temporal Memory Attention for Video Semantic
  Segmentation}.
\newblock \bibinfo{journal}{\emph{arXiv preprint arXiv:2102.08643}}
  (\bibinfo{year}{2021}).
\newblock


\bibitem[\protect\citeauthoryear{Wang, Zhou, Yan, and Deng}{Wang
  et~al\mbox{.}}{2018}]%
        {wang2018fully}
\bibfield{author}{\bibinfo{person}{Shiyao Wang}, \bibinfo{person}{Yucong Zhou},
  \bibinfo{person}{Junjie Yan}, {and} \bibinfo{person}{Zhidong Deng}.}
  \bibinfo{year}{2018}\natexlab{}.
\newblock \showarticletitle{Fully motion-aware network for video object
  detection}. In \bibinfo{booktitle}{\emph{Proceedings of the European
  conference on computer vision (ECCV)}}. \bibinfo{pages}{542--557}.
\newblock


\bibitem[\protect\citeauthoryear{Wang, Chao, Garg, Hariharan, Campbell, and
  Weinberger}{Wang et~al\mbox{.}}{2019}]%
        {wang2019pseudo}
\bibfield{author}{\bibinfo{person}{Yan Wang}, \bibinfo{person}{Wei-Lun Chao},
  \bibinfo{person}{Divyansh Garg}, \bibinfo{person}{Bharath Hariharan},
  \bibinfo{person}{Mark Campbell}, {and} \bibinfo{person}{Kilian~Q
  Weinberger}.} \bibinfo{year}{2019}\natexlab{}.
\newblock \showarticletitle{Pseudo-lidar from visual depth estimation: Bridging
  the gap in 3d object detection for autonomous driving}. In
  \bibinfo{booktitle}{\emph{Proceedings of the IEEE/CVF Conference on Computer
  Vision and Pattern Recognition}}. \bibinfo{pages}{8445--8453}.
\newblock


\bibitem[\protect\citeauthoryear{Wang, Jiang, Gao, Hwang, Xing, and Liu}{Wang
  et~al\mbox{.}}{2021a}]%
        {wang2021rodnet}
\bibfield{author}{\bibinfo{person}{Yizhou Wang}, \bibinfo{person}{Zhongyu
  Jiang}, \bibinfo{person}{Xiangyu Gao}, \bibinfo{person}{Jenq-Neng Hwang},
  \bibinfo{person}{Guanbin Xing}, {and} \bibinfo{person}{Hui Liu}.}
  \bibinfo{year}{2021}\natexlab{a}.
\newblock \showarticletitle{Rodnet: Radar object detection using cross-modal
  supervision}. In \bibinfo{booktitle}{\emph{WACV}}. \bibinfo{pages}{504--513}.
\newblock


\bibitem[\protect\citeauthoryear{Yue, Wu, Seshia, Keutzer, and
  Sangiovanni-Vincentelli}{Yue et~al\mbox{.}}{2018}]%
        {yue2018lidar}
\bibfield{author}{\bibinfo{person}{Xiangyu Yue}, \bibinfo{person}{Bichen Wu},
  \bibinfo{person}{Sanjit~A Seshia}, \bibinfo{person}{Kurt Keutzer}, {and}
  \bibinfo{person}{Alberto~L Sangiovanni-Vincentelli}.}
  \bibinfo{year}{2018}\natexlab{}.
\newblock \showarticletitle{A lidar point cloud generator: from a virtual world
  to autonomous driving}. In \bibinfo{booktitle}{\emph{Proceedings of the 2018
  ACM on International Conference on Multimedia Retrieval}}.
  \bibinfo{pages}{458--464}.
\newblock


\bibitem[\protect\citeauthoryear{Yun, Han, Oh, Chun, Choe, and Yoo}{Yun
  et~al\mbox{.}}{2019}]%
        {yun2019cutmix}
\bibfield{author}{\bibinfo{person}{Sangdoo Yun}, \bibinfo{person}{Dongyoon
  Han}, \bibinfo{person}{Seong~Joon Oh}, \bibinfo{person}{Sanghyuk Chun},
  \bibinfo{person}{Junsuk Choe}, {and} \bibinfo{person}{Youngjoon Yoo}.}
  \bibinfo{year}{2019}\natexlab{}.
\newblock \showarticletitle{Cutmix: Regularization strategy to train strong
  classifiers with localizable features}. In
  \bibinfo{booktitle}{\emph{Proceedings of the IEEE/CVF International
  Conference on Computer Vision}}. \bibinfo{pages}{6023--6032}.
\newblock


\bibitem[\protect\citeauthoryear{Yun, Oh, Heo, Han, and Kim}{Yun
  et~al\mbox{.}}{2020}]%
        {yun2020videomix}
\bibfield{author}{\bibinfo{person}{Sangdoo Yun}, \bibinfo{person}{Seong~Joon
  Oh}, \bibinfo{person}{Byeongho Heo}, \bibinfo{person}{Dongyoon Han}, {and}
  \bibinfo{person}{Jinhyung Kim}.} \bibinfo{year}{2020}\natexlab{}.
\newblock \showarticletitle{VideoMix: Rethinking Data Augmentation for Video
  Classification}.
\newblock \bibinfo{journal}{\emph{arXiv preprint arXiv:2012.03457}}
  (\bibinfo{year}{2020}).
\newblock


\bibitem[\protect\citeauthoryear{Zhao, Li, Abu~Alsheikh, Tian, Zhao, Torralba,
  and Katabi}{Zhao et~al\mbox{.}}{2018}]%
        {zhao2018through}
\bibfield{author}{\bibinfo{person}{Mingmin Zhao}, \bibinfo{person}{Tianhong
  Li}, \bibinfo{person}{Mohammad Abu~Alsheikh}, \bibinfo{person}{Yonglong
  Tian}, \bibinfo{person}{Hang Zhao}, \bibinfo{person}{Antonio Torralba}, {and}
  \bibinfo{person}{Dina Katabi}.} \bibinfo{year}{2018}\natexlab{}.
\newblock \showarticletitle{Through-wall human pose estimation using radio
  signals}. In \bibinfo{booktitle}{\emph{Proceedings of the IEEE Conference on
  Computer Vision and Pattern Recognition}}. \bibinfo{pages}{7356--7365}.
\newblock


\bibitem[\protect\citeauthoryear{Zhu, Wang, Dai, Yuan, and Wei}{Zhu
  et~al\mbox{.}}{2017}]%
        {zhu2017flow}
\bibfield{author}{\bibinfo{person}{Xizhou Zhu}, \bibinfo{person}{Yujie Wang},
  \bibinfo{person}{Jifeng Dai}, \bibinfo{person}{Lu Yuan}, {and}
  \bibinfo{person}{Yichen Wei}.} \bibinfo{year}{2017}\natexlab{}.
\newblock \showarticletitle{Flow-guided feature aggregation for video object
  detection}. In \bibinfo{booktitle}{\emph{Proceedings of the IEEE
  International Conference on Computer Vision}}. \bibinfo{pages}{408--417}.
\newblock


\end{thebibliography}




\end{document}